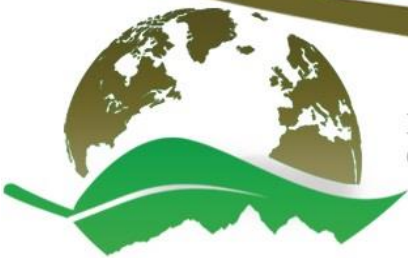

# Optimal Image Smoothing and Its Applications in Anomaly Detection in Remote Sensing

M. Kiani

*Abstract*—This paper is focused on deriving an optimal image smoother. The optimization is done through the minimization of the norm of the Laplace operator in the image coordinate system. Discretizing the Laplace operator and using the method of Euler-Lagrange result in a weighted average scheme for the optimal smoother. Satellite imagery can be smoothed by this optimal smoother. It is also very fast and can be used for detecting the anomalies in the image. A real anomaly detecting problem is considered for the Qom region in Iran. Satellite image in different bands are smoothed. Comparing the smoothed and original images in different bands, the maps of anomalies are presented. Comparison between the derived method and the existing methods reveals that it is more efficient in detecting anomalies in the region.

## I. INTRODUCTION

THE methods of smoothing for satellite images in the field of remote sensing have always been needed. One fundamental reason behind this demand is that almost all images have a certain degree of noise, due to the environmental and satellites' imaging instrument. Realizing this fact, there have been many studies on the smoothing techniques and their improvement. These include, but are not confined to, [1]-[8]. In [1], a fast image smoothing algorithm is proposed, which is based on the Gaussian distribution and its probability characteristics. In this method, only those pixels that are within a certain sigma probability (in the Gaussian sense) are used for smoothing, in a weighted average scheme. The powerful characteristic of this method is its ability to save the nuances in the image, while smoothing it at the same time. This means the details in the image are not destroyed by the averaging process. This method has various applications in the field of image processing, including contrast enhancing and image segmentation. In [2], in contrast to the method described in [1], the smoothing method is designed in a way such that it makes the edges more pronounced. The method is based on the global $L^0$ gradient minimization, which removes most of the less dramatic changes in the image. Since it is a global approach, it cannot be used for image features controlling in a local manner. However, as it was stated earlier, it is a powerful means to extract edges, an important application in the field of remote sensing and photogrammetry. In [3], a texture-based method is presented, which is based on the covariances in a particular region in the image. It is an image processing tool to extract the so-called structures in the image from the texture of the image. Its usefulness can be understood if one wants to decompose an image and extract some pre-intended characteristics of the image. [4] presents a rational-approximation approach for the noise reduction in an image. This method works for a variety of noise distributions. Preservation of the image's essential details is one of the most important characteristics of this method. The image smoothing method presented in [5] is based on the graph theory. Considering the image to be smoothed as a graph, a mean curvature smoothing method is derived and its efficiency in image enhancement is demonstrated in both gray-level and color images. In [6] a global method is proposed for image smoothing that is based on the so-called Laplacian matrices. Its powerful characteristic-edge preservation-is combined with the smoothing optimization, thus making it an ideal tool for many applications in the field of image processing. The proposed image smoothing tool in [7] is based on the heat kernels and their associated Laplacian matrices. The method smoothes the image but does not make it blurred. The implementational procedure is carried out in the standard convolution manner, i.e., a numerical kernel is convoluted with the image to give the smoothed value of each individual pixel in the image. In [8], an image smoothing scheme is presented by which the image can be smoothed various times. In other words, this method is a multi-resolution approach. The smoothing works well in both the continuous and discontinuous regions in the image. In the latter case, a weighted least squared approach is used to tackle the discontinuity problem. An important feature of this method is that it can be used for the SAR images in the field of interferometry in remote sensing.

Smoothing techniques have widespread use in different areas, from the recognition of the faces to sharpening and defining

Manuscript received January 28, 2020: Revised version received January 28, 2020. This work is not done under the support of any institution or person.
  M. Kiani was with the University of Tehran, Enghelab Square, Tehran, Iran (phone: +98-910-0035865; e-mail: mostafakiani@ut.ac.ir).
.

higher resolution images, and to extraction of essential and vital information from the images. One of the most important fields in which the image processing techniques are crucial tools to implement the required, further procedures is remote sensing. Smoothing techniques are used for a variety of tasks in this field, including removing the effect of satellite's instrument and atmospheric errors. One of the most important objectives in remote sensing is detecting some unusual phenomena, things that are significantly (or sometimes subtly) different from the main textural elements in the image. An example of this kind of phenomena is a gold mine, which can be detected from satellite imagery, using the corresponding pixels' physical characteristics that are detectable in the image, including intensity, radiance, and reflectance. There are many references that deal with this so-called anomaly detection problem, including [9]-[11].

In [9], satellite hyperspectral imagery is the basis for detecting anomalies in the region represented in the image. Since the hyperspectral imagery is based on an image in different wavelengths, it allows us to detect an anomaly much more confidently than other types of imagery, like the multispectral approaches. The method described in this paper is a simultaneous anomaly detection and classification approach, resolving the issues of undistinguishable anomalies. The method described in [10] is primarily useful for crop yield predictions. Since the images used for agricultural purposes (especially crop yield) are typically low-resolution, a distinct method of yield anomaly detection is presented to deal with this problem. In [11], a method is devised to detect anomalies in the hyperspectral imagery. The method is based on the assumption that each regular pixel in the image can be represented as a linear combination of its surrounding neighbors, carried out in a $L^2$ norm minimization. In contrast to the regular pixels in the image, the anomalous ones cannot be represented in the mentioned form, thus being distinguishable. The method we present in this paper is based on this fundamental principle. However, we perform $L^2$ semi-norm minimization for the Laplacian operator. This is performed in a similar way to [12] and [13]. This is done to derive a global smoother, which can be used for detecting anomalies as well.

The rest of this paper is organized as follows. In section 2, the derivation of the optimal image smoother is presented. Section 3 is devoted to the remote sensing application of the theoretical results obtained in section 2. Final notes and conclusions are stated in section 4.

## I. DERIVATION OF THE OPTIMAL IMAGE SMOOTHING

In this section the theoretical discussion of the optimal image smoothing is presented.

It is well known that the changes in a function can be characterized by differential operators. Among these differential operators is the Laplacian, a second order, elliptic differential operator whose norm minimization corresponds to the smoothest function in a given space (see [12], [14], and [15] for more details.) This fundamental assumption is crucial for the anomaly detection application: like the method used in [11], the anomalous pixels cannot be written in a linear form the coefficients of which are determined based on the neighboring pixels' values. However, it is important to notice that here the smoothest solution is found, contrary to [11]. Thus, it can be said that the method presented in this paper is optimal in the sense of image smoothing.

### A. Mathematical background of the optimization

To present the image smoothing optimization definition and solution, it is necessary to give some background about the problem.

The Laplace operator, denoted hereafter by $\Delta$, in the two-dimensional flat space whose horizontal and vertical axes are, respectively, $x$ and $y$, is defined as the following

$$\Delta = \frac{\partial^2}{\partial x^2} + \frac{\partial^2}{\partial y^2}. \qquad (1)$$

The set of all infinitely often differentiable functions in the domain D, whose Laplacians are square-integrable is a semi-Hilbert space denoted as $H$ (see [12] and [13] for more details)

$$H(D) = \{f \mid f, \Delta f \in L^2(D)\}. \qquad (2)$$

The norm in the relation (2) is in the sense of Lebesgue integral. One of the most important questions that might arise is whether there exists a function in the space in (2) that has the highest degree of smoothness. The answer to the question above is positive (see [14] and [15].) It is even interesting to notice that this function is unique ([12], [14], and [15].) Based on the norm in (2), one can define the definition of the smoothest function. The function $S \in H(D)$ is called the smoothest function in $H(D)$ if

$$S(x,y) = \arg(\min\|\Delta f\|^2), f \in H(D). \qquad (3)$$

The relation in (3) is equivalent to

$$\iint (\Delta f)^2 dx dy \to min. \qquad (4)$$

Usually the function $S$ is called a two-dimensional spline.

### B. The solution of the minimization problem

The solution of the relation in (4) gives us the continuous spline function. However, as the primary purpose of this paper is smoothing of an image, we need to pursue the problem in a discrete perspective. In a similar way as the ones described in [12] and [13], the Laplacian and the function itself are discretized. Besides, the method of Euler-Lagrange is carried out for the minimization. Hence, regarding the methods used in [12] and [13], we perform the minimization in (4) in distinct steps, as follows.

1. The following norm is to be minimized

$$Q(f) = \iint (\Delta f)^2 dx dy. \qquad (5)$$

Thus, the Euler-Lagrange method, with the function $g$ in conjunction with the function $f$, gives

$$dQ(f,g) = \lim_{\epsilon \to 0} \frac{Q(f+\epsilon g) - Q(f)}{\epsilon}. \quad (6)$$

Using the definition of the operator $Q$, given in (5), in (6), one gets

$$dQ(f,g) = \iint \Delta f \Delta g \, dxdy. \quad (7)$$

The Euler-Lagrange method forces the following equation to hold

$$dQ(f,g) = 0. \quad (8)$$

To find the equivalent condition on $f$ in (7), one needs to have the eigenvalues and eigenfunctions of the Laplace operator (see [12].) This is a simple problem and can be found in many different references, including [16]. However, we only use the symbols of these values and functions. The symbols are, respectively, n and $h_{n,m}$. Thus, we have

$$\Delta h_{n,m}(x,y) = n \, h_{n,m}(x,y). \quad (9)$$

The index $n$ in $h_n$ means the $n^{th}$ eigenfunction, and $m$ represents its order. This is because corresponding to each eigenvalue there exists $m$ eigenfunctions. Besides, these functions are orthogonal pairwise, meaning

$$\iint h_{n,m}(x,y) h_{k,l}(x,y) dxdy = 0, \text{ if } n \neq k \text{ or } m \neq l. \quad (10)$$

Hence, one can simply expand $f$ and $g$ in (7), based on the eigenfunctions in (9)

$$f = \sum_{n,m} a_{n,m} h_{n,m},$$
$$g = \sum_{k,l} b_{k,l} h_{l,k}, \quad (11)$$

where $a_{n,m}$ and $b_{k,l}$ are the coefficients of the expansion.
Inserting the relations (11) in (7) and considering the condition in (8), one gets

$$\iint \Delta(\sum_{n,m} a_{n,m} h_{n,m}) \Delta(\sum_{k,l} b_{k,l} h_{k,l}) \, dxdy = 0. \quad (12)$$

Using the relation in (9) and considering the orthogonality of the eigenfunctions in (10), one arrives at the following relation

$$\iint \Delta^2 f \, g \, dxdy = 0. \quad (13)$$

The equation in (13) holds for any arbitrary choice of function $g$ if and only if

$$\Delta^2 f = 0. \quad (14)$$

(14) is normally called "biharmonic" equation.

2. The next step is to discretize the Laplacian and its biharmonic form in (14).

For this purpose, one must write the following form for the Laplacian operator

$$\Delta f(x_i, y_j) = \sum_{p=-1}^{+1} \sum_{q=-1}^{+1} d_{p,q} f(x_{i+p}, y_{j+q}), \quad (15)$$

in which $i$ and $j$ are the $i^{th}$ and $j^{th}$ $x$ and $y$, respectively, and the coefficients $d_{p,q}$ are determined by the Taylor expansion of the function $f$. However, we do not need these values. Rather, we need to insert (15) in (14), and get the following relation and its corresponding coefficients $c_{p,q}$

$$\Delta^2 f(x_i, y_j) = \sum_{p=-2}^{+2} \sum_{q=-2}^{+2} c_{p,q} f(x_{i+p}, y_{j+q}). \quad (16)$$

It is essential to notice that the increments in $x$ and $y$ directions are, accordingly, $l_x$ and $l_y$. This means we have

$$x_{i+p} - x_i = pl_x,$$
$$y_{j+q} - y_j = ql_y. \quad (17)$$

In order to calculate the 25 coefficients $c_{p,q}$, we expand the function $f$ around the point $(x_i, y_j)$

$$f(x_{i+p}, y_{j+q}) = f(x_i, y_j) + pl_x \frac{\partial f}{\partial x} + ql_y \frac{\partial f}{\partial y} + \frac{p^2 l_x^2}{2} \frac{\partial^2 f}{\partial x^2}$$
$$+ \frac{q^2 l_y^2}{2} \frac{\partial^2 f}{\partial y^2} + pq l_x l_y \frac{\partial^2 f}{\partial x \partial y} + \cdots. \quad (18)$$

Since there are 25 unknown coefficients in (16), the Taylor expansion of the function $f$ in (18) must contain 25 terms for the mentioned coefficients to be calculated. This is done by expanding the $f$ up to degree 6 ($\frac{\partial^6 f}{\partial x^u \partial y^v}, u + v = 6$), in addition to the first for terms of the seventh degree. After this, the coefficients $c_{p,q}$ in (16) are calculated, using the equations below

$$\Delta^2 f(x_i, y_j) = \Delta(\Delta f|_{x_i, y_j}) = \Delta\left(\frac{\partial^2 f}{\partial x^2}\bigg|_{x_i, y_j} + \frac{\partial^2 f}{\partial y^2}\bigg|_{x_i, y_j}\right)$$
$$= \frac{\partial^4 f}{\partial x^4}\bigg|_{x_i, y_j} + 2\frac{\partial^4 f}{\partial x^2 \partial y^2}\bigg|_{x_i, y_j} + \frac{\partial^4 f}{\partial y^4}\bigg|_{x_i, y_j}, \quad (19)$$

$$\sum_{p=-2}^{+2} \sum_{q=-2}^{+2} c_{p,q} f(x_{i+p}, y_{j+q}) = \frac{\partial^4 f}{\partial x^4} + \frac{\partial^4 f}{\partial y^4} + 2\frac{\partial^4 f}{\partial x^2 \partial y^2}. \quad (20)$$

From (18) and (20), the coefficients $c_{p,q}$ are as the following

$$c_{-2,-2} = c_{-2,-1} = c_{-2,1} = c_{-2,+2} = c_{-1,-2} = c_{-1,+2} = c_{+1,-2}$$
$$= c_{+1,+2} = c_{+2,-2} = c_{+2,-1} = c_{+2,+1} = c_{+2,+2} = 0, \quad (21)$$

$$c_{-2,0} = c_{2,0} = \frac{1}{l_x^4}, \quad (22)$$

$$c_{-1,-1} = c_{-1,+1} = c_{+1,-1} = c_{+1,+1} = \frac{2}{l_x^2 l_y^2}, \quad (23)$$

$$c_{-1,0} = c_{+1,0} = -4\frac{l_x^2 + l_y^2}{l_x^4 l_y^2}, \quad (24)$$

$$c_{0,-2} = c_{0,+2} = \frac{1}{l_y^4}, \quad (25)$$

$$c_{0,-1} = c_{0,+1} = -4\frac{l_x^2 + l_y^2}{l_x^2 l_y^4}, \quad (26)$$

$$c_{0,0} = 2\frac{3l_x^4 + 3l_y^4 + 4l_x^2 l_y^2}{l_x^4 l_y^4}. \quad (27)$$

3. The final step is to derive an explicit template with which the image can be smoothed.

In (21)-(27), if the increments are defined based on the pixel increment, i.e. 1, the following template is derived, shown in Fig 1.

|   |   |   |   |   |
|---|---|---|---|---|
| 0 | 0 | 1 | 0 | 0 |
| 0 | 2 | -8 | 2 | 0 |
| 1 | -8 | 20 | -8 | 1 |
| 0 | 2 | -8 | 2 | 0 |
| 0 | 0 | 1 | 0 | 0 |

*Figure 1: The optimal image smoothing template*

It is important to notice that the template shown in Fig 1 is a special kind of the more general class of templates defined based on the various possible values for $l_x$ and $l_y$. However, it is the simplest template that can be defined and thus, we have used this template to smooth the satellite imagery in the next section.

## II. APPLICATION OF THE OPTIMAL IMAGE SMOOTHING IN THE FIELD OF REMOTE SENSING: ANOMALY DETECTION IN THE QOM REGION IN IRAN

In this section a possible application of the derived template in the previous section is presented. One of the most important fields in which satellite imagery plays the paramount role is the field of remote sensing. In this field, by using the physical laws of the quantifiable characteristics of the phenomena in the image, and by the methods of image processing, information about the things in the image is gained. One important piece of information that can be acquired from an image is finding the so-called anomalous phenomena, the characteristics of which are different from those of the main texture of the image. An example of an anomaly is a mine in a region, whose corresponding pixels' spectral signature is different from that of the surrounding pixels.

Since the anomalies in the image have different characteristics from the main pixels in the image, they cannot be represented based on the main pixels. This means if the image is smoothed, anomalous pixels are evident in the image. This is the fundamental assumption behind many methods of detecting anomalies from the satellite images. Based on this principle, we present a case study of anomaly detection by the optimal image smoother, derived in the previous section. This case study is for the Qom region in Iran. The most important features of this region are the Salt Lake, located in northern part of this region, and the Qom city, in the southern part of it. Our method is to distinguish these two features from the usually arid nearby regions. For this reason, we have used Landsat satellite imagery in 8 bands.

Fig 2 shows the Qom region.

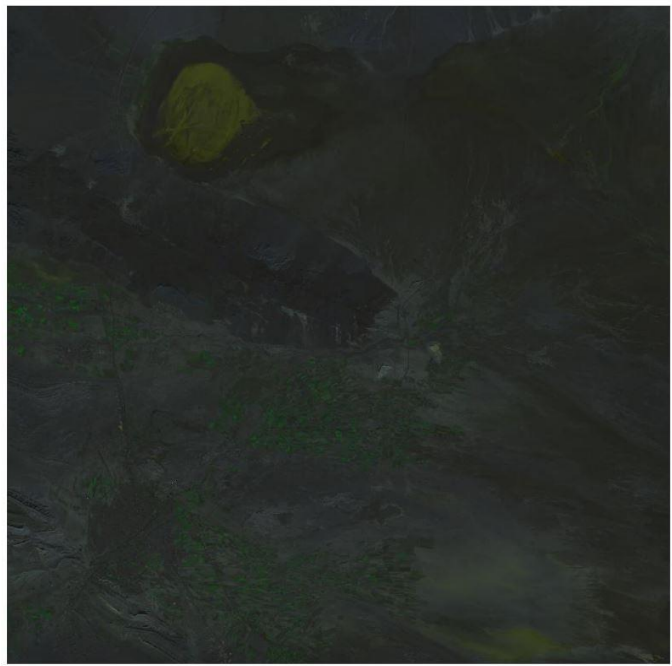

*Figure 2: Qom region in Iran, in Landsat satellite imagery*

Applying the optimal image smoothing template in Fig 1 to the image in all of its 8 bands and then subtracting the unsmoothed images from the corresponding smoothed ones, we get the following images

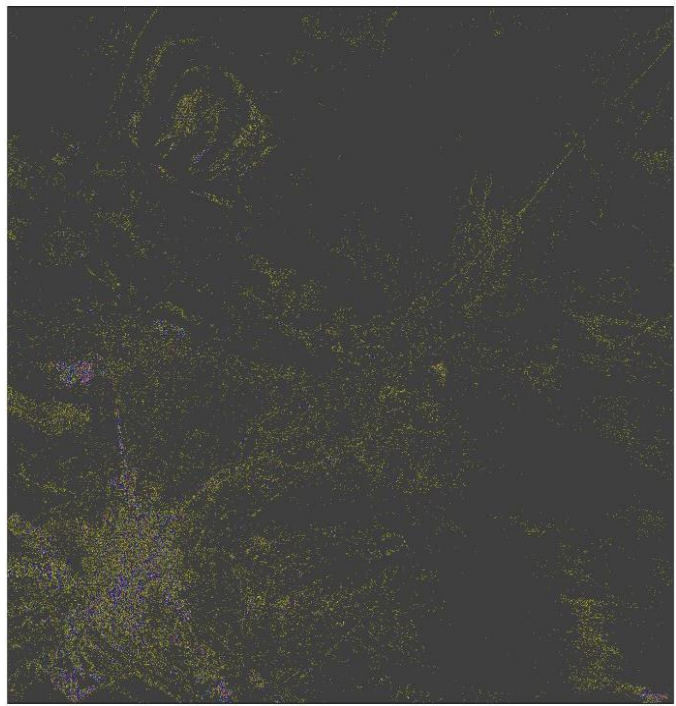

*Figure 3: Anomalies in band 1*

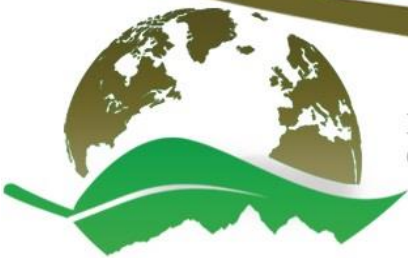

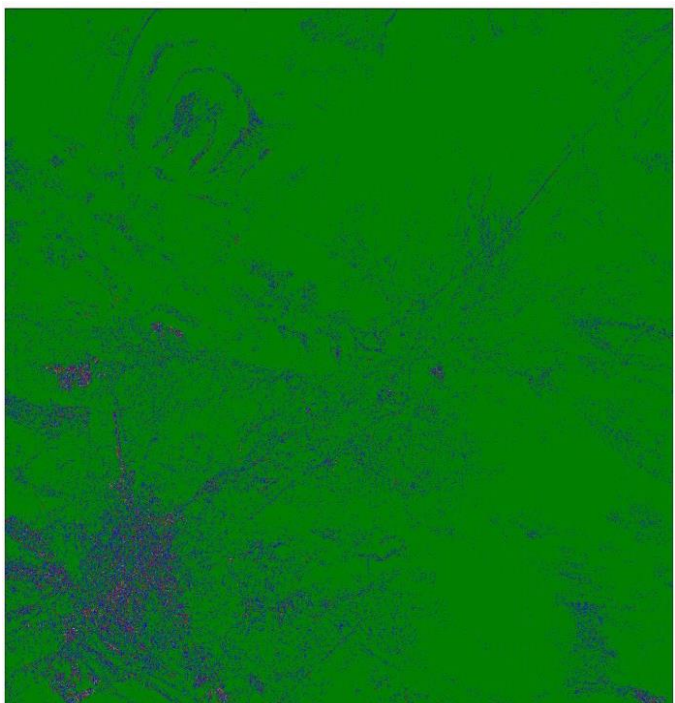

*Figure 4: Anomalies in band 2*

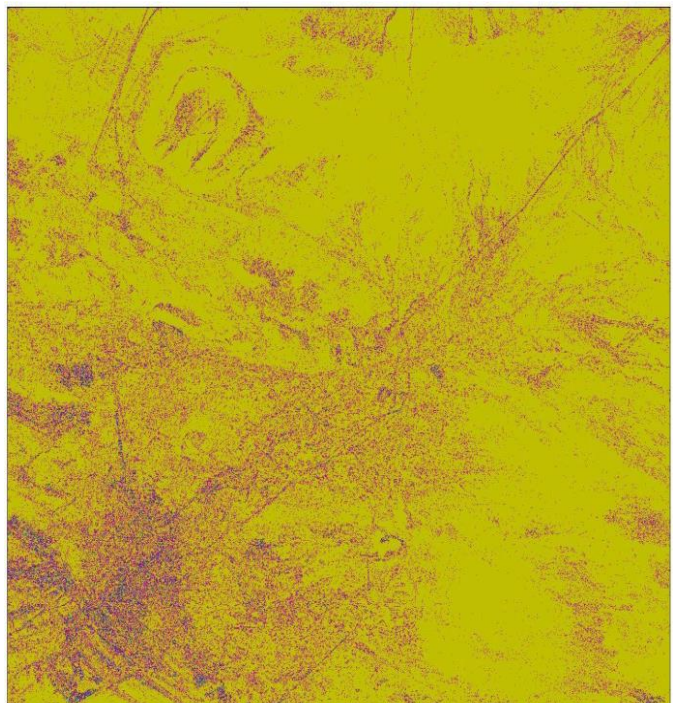

*Figure 6: Anomalies in band 4*

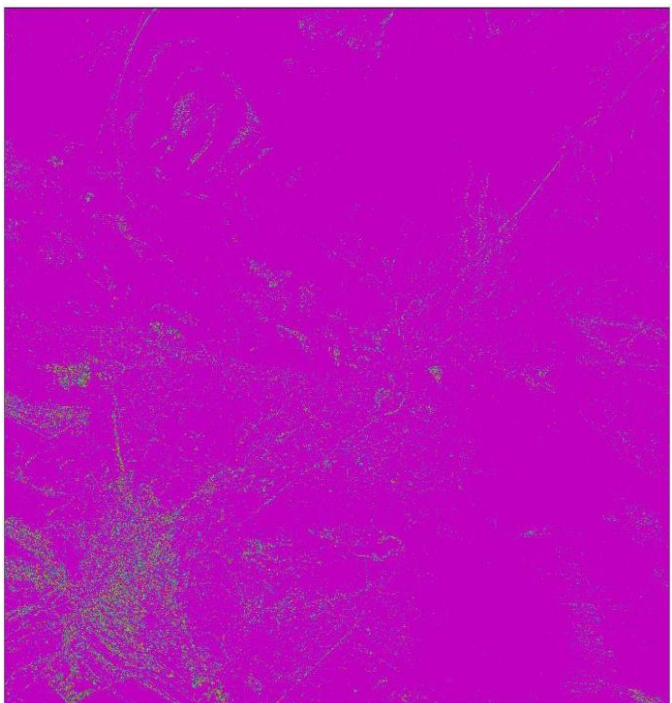

*Figure 5: Anomalies in band 3*

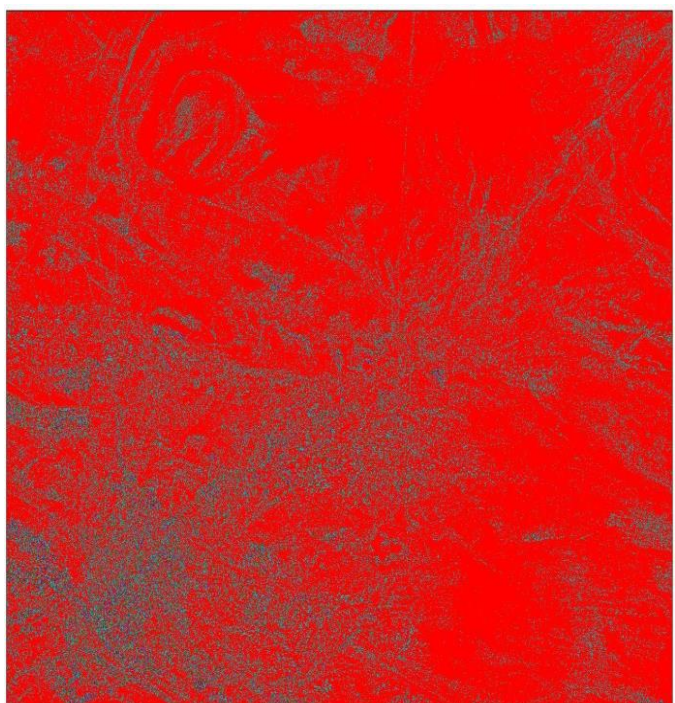

*Figure 7: Anomalies in band 5*

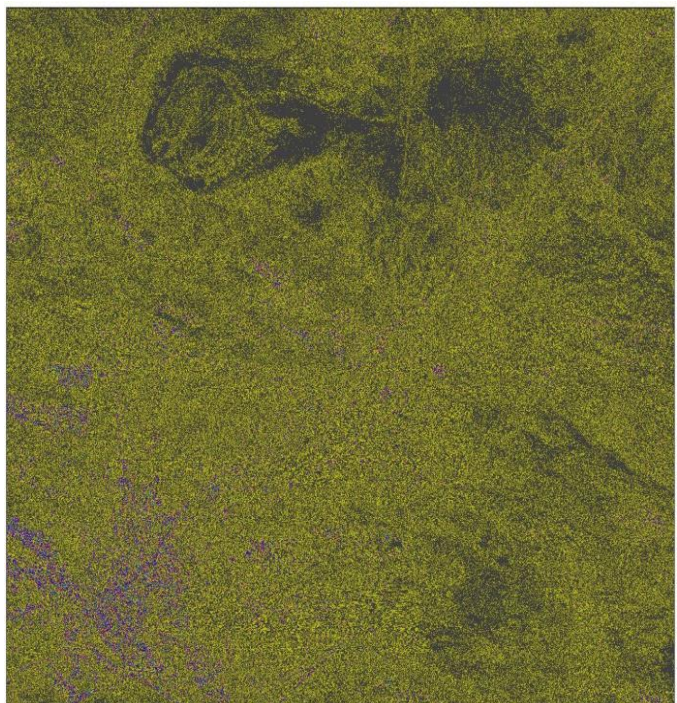

*Figure 8: Anomalies in band 6*

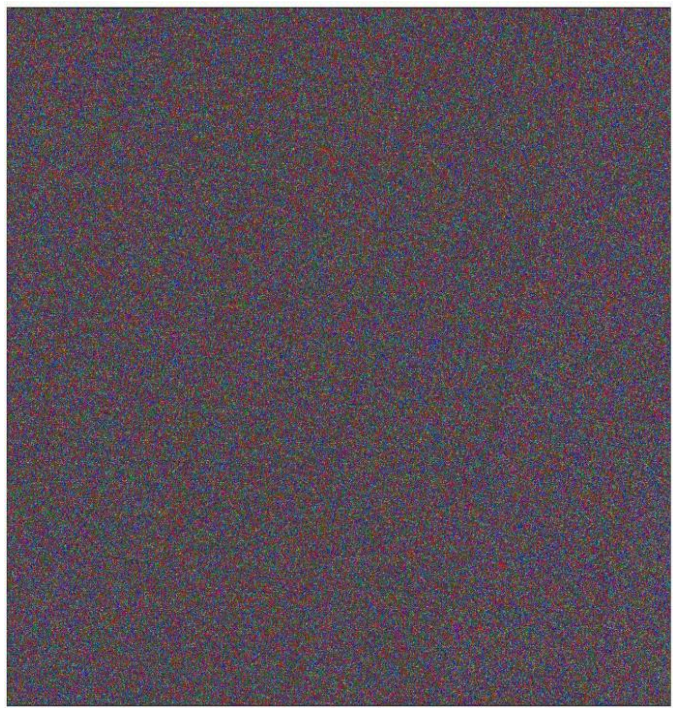

*Figure 10: Anomalies in band 8*

Some important points about the Figs 3-10 are

1. As the band number increases from 1 to 8 the anomalies are less detectable. This is because the wavelength in band 8 is larger, so the difference between adjacent points are less evident. In other words, the higher the band's number, the more uniform the pixel values become.
2. The derived method in this paper is more efficient in detecting anomalies. To show this, we have used one common high-pass-filtering for anomaly detection. This filter is shown in the following, which is normally called 2D Laplacian.

| -1 | -1 | -1 |
|----|----|----|
| -1 | 8  | -1 |
| -1 | -1 | -1 |

*Figure 11: Two-dimensional Laplacian for anomaly detection*

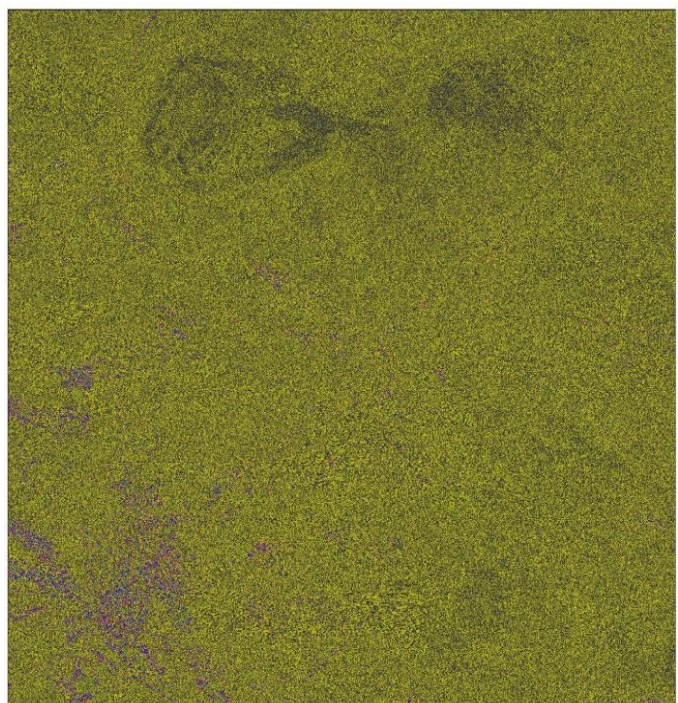

*Figure 9: Anomalies in band 7*

In Fig 12, the anomalies detected by this filter are shown in just band 1.

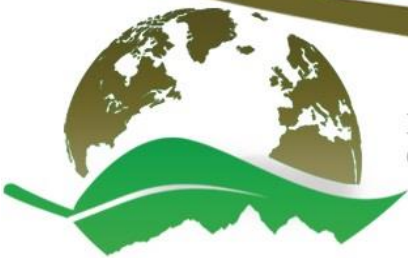

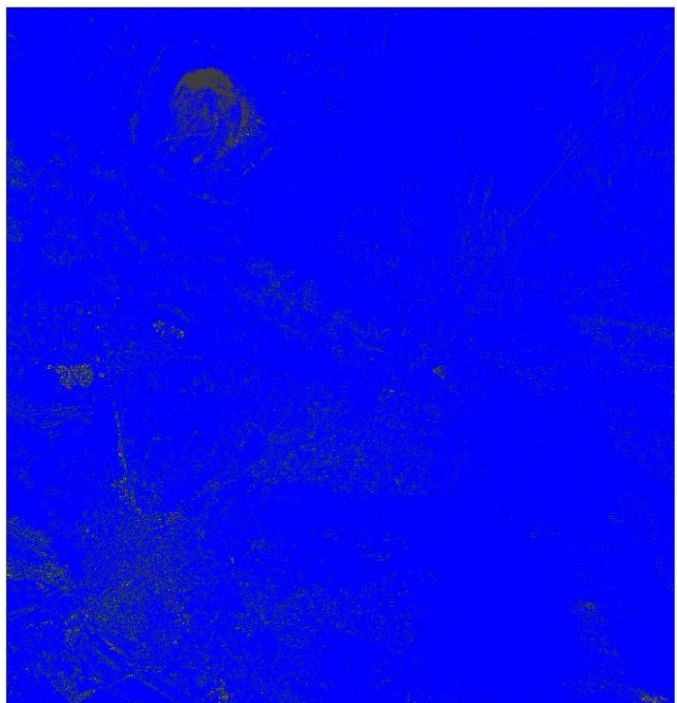

*Figure 12: Anomalies by the Laplacian method*

Comparing the Figs 3 and 12, one can understand that the method presented in this paper is better for detecting anomalies.

3. In Fig 13, the classification of the anomalies is presented, using parallelepiped scheme in 8 bands, with the overall accuracy of 0.7325.

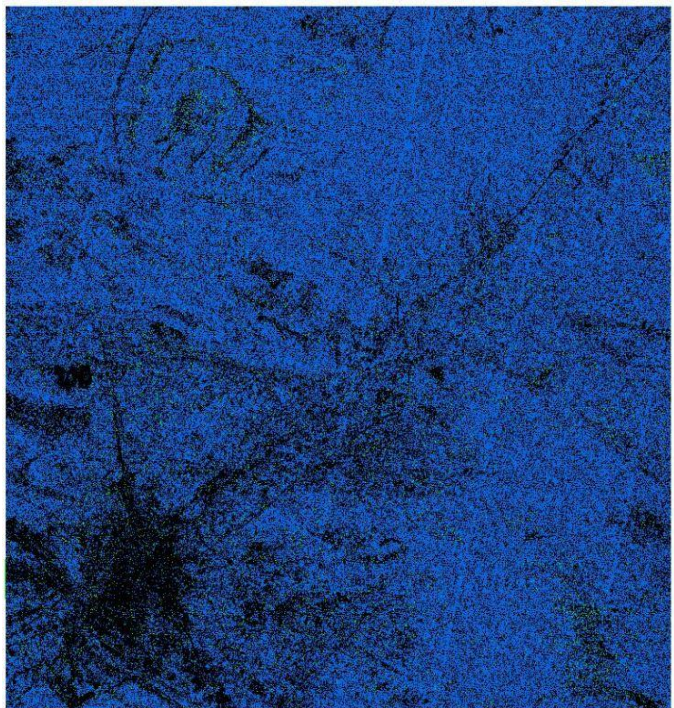

*Figure 13: Classification of anomalies detected by the new method*

### III. CONCLUSIONS

In the present paper a new method for image smoothing is presented, which is based on the $L^2$ semi-norm minimization, done by minimizing the Laplace operator in the two-dimensional, discrete image space. The method results in a $5 \times 5$ template, which must be convoluted with the original image. Subtraction of the original image from the smoothed image would show anomalies. A case study is presented for the Qom region in Iran, in which the Qom city and its Salt Lake are detected in the image. It is shown that this method is more efficient than the common methods, such as the two-dimensional Laplacian. The method used in this paper can be a starting point for many future works on the anomaly detection problem in the field of remote sensing. One approach can be the unification of this method with other successful methods such as the one in [11]. This can be done in a multi-resolution analysis. Future works may focus on this area for more accurate anomaly detectors.

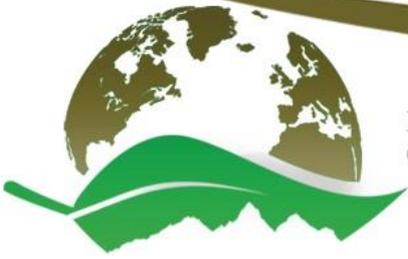

M. Kiani is an all-round Geoscientist who works in all areas of Geoscience, especially in Geophysics and Geodesy. He works in both theoretical and applied aspects of all areas of Geosciences, among which are gravity field, physical, satellite, and mathematical Geodesy, Remote Sensing, and estimation theory. He was graduated from University of Tehran in 2018. His research in all areas of Geosciences is ongoing.